\documentclass[a4paper]{article}

\usepackage{CJKutf8}
\usepackage{hyperref, url}
\usepackage{amsmath,graphicx}
\usepackage{INTERSPEECH2019}

\title{Code-switching Sentence Generation by Generative Adversarial \\Networks and its Application to Data Augmentation}
\name{Ching-Ting Chang, Shun-Po Chuang, Hung-Yi Lee}
\address{Graduate Institute of Communication Engineering, National Taiwan University}
\email{r05942066@ntu.edu.tw, f04942141@ntu.edu.tw, hungyilee@ntu.edu.tw}

\begin{document}

\maketitle
\begin{abstract}
  Code-switching is about dealing with alternative languages in speech or text. It is partially speaker-dependent and domain-related, so completely explaining the phenomenon by linguistic rules is challenging. Compared to most monolingual tasks, insufficient data is an issue for code-switching. To mitigate the issue without expensive human annotation, we proposed an unsupervised method for code-switching data augmentation. By utilizing a generative adversarial network, we can generate intra-sentential code-switching sentences from monolingual sentences. We applied the proposed method on two corpora, and the result shows that the generated code-switching sentences improve the performance of code-switching language models.
\end{abstract}

\section{Introduction}
\label{sec:intro}
Code-switching (CS) is the practice that two or more languages are used within a document or a sentence. It is widely observed in multicultural areas, or countries where official language is different from native language. For example, Taiwanese tend to mix English and Taiwanese Hokkien in their text and speech besides their main language, Mandarin. Solving CS is crucial to building a general ASR system that can process both monolingual and CS speech~\cite{ccetinouglu2016challenges, zeng2017improving, xu2018pruned}.  In this paper, we focus on improving the language models for ASR of intra-sentential CS speech.
Specifically, we only deal with words and phrases that are code-switched within a sentence.

Computational processing of CS is fundamentally challenging due to lack of data. Applying linguistic knowledge is a solution to this~\cite{bhatt1995code,pfaff1979constraints}. Equivalence Constraint and Functional Head Constraint are used to build a better CS language model~\cite{li2012code,li2013improved,ying2014language}, and CS models with syntactic and semantic features are built to exploit more information~\cite{adel2015syntactic,yeh2015improved}. Because of a large amount of monolingual data, monolingual language models for host and guest languages are learned separately, and then combined with a probabilistic model for switching between the two~\cite{garg2017dual}.

Because CS is mostly used in spoken language, the most practical way of generating data is to label CS speech. However, manual transcription requires plenty of skilled labor and hours of tedious work.  An alternative way is to generate CS data from existing monolingual text. Unfortunately, there are no flawless rules for predicting code-switching points within a sentence, since each person tends to code-switch in a different manner. These years, people try to synthesize more code-switching text by the models learned from data~\cite{yilmaz2018acoustic,garg2018code, yilmaz2018building}.

Generative models have been used to generate CS sentences~\cite{garg2018code}, but previous work uses generative model to generate the sentences from scratch. Here the generator learns to modify monolingual sentences into CS sentences. In this way, the generator can leverage the information from monolingual sentences.

We propose a novel CS text generation method, by using generative adversarial networks (GAN)~\cite{goodfellow2014generative} with reinforcement learning (RL)~\cite{yu2017seqgan}, to generate CS data from monolingual sentences automatically. With CS data augmented by our method, it is possible to solve the problem of sparse training data. Our proposed method has the following benefits:
\begin{itemize}
\itemsep -2pt
\item We don't use any labeled data to train the generator.
\item The model learns CS rules for data generation implicitly with the help of discriminator instead of defining hand-crafted rules.
\item We conduct the experiments on two Mandarin-English code-switching  corpora, LectureSS and SEAME, which have very different characteristics to show that the proposed approach generalizes well in different cases.
\end{itemize}

The experimental results show that GAN can generate reasonable code-switching sentences, and the generated  code-switching sentences can be used to improve language modeling.

\section{Methodology}
\label{sec:method}

\begin{table*}[ht]
\caption{\footnotesize Details for the corpora: LecureSS and SEAME.}
\label{table:corpus}
\centering {\footnotesize
\begin{tabular}{l|ccc|ccc}
\hline
 & \multicolumn{3}{c}{LectureSS} & \multicolumn{3}{c}{SEAME} \\
\cline{2-7}
 & {train} & {dev} & {test} \\
 \hline
{\# speakers} & - & - & - & 139 & 8  & 8 \\
{\# total utterances} & 12657 & 2634 & 1500 & 94055 & 6115 & 5908 \\
{\# Mandarin utterances} & 4643 & 964 & 810  & 20365 & 1653 & 2211 \\
{\# code-switching utterances} & 8014 & 1670 & 690 & 50421 & 3564 & 3066 \\
{\# total words} & 160862 & 28063 & 12398  & 964927 & 65696 & 57620 \\
{\# Mandarin words} & 138409 & 23308  & 10871 & 543399 & 42525 & 41874 \\
{\# English words} & 22453 & 4755 & 1527 & 417452 & 22641 & 15370 \\
\hline
\end{tabular} }
\end{table*}
\begin{table}[ht]
\caption{\footnotesize Comparison between LectureSS and SEAME.}
\label{table:corpusvs}
\centering {\footnotesize
\begin{tabular}{l|c|c}
\hline
 & {LectureSS} & {SEAME}\\
\hline
{\# speakers}    & 1 & 156  \\
{Nationality} & Taiwan & Singapore \& Malaysia  \\
{Domain}      & signal lecture & daily, school \\
{ZH:EN words} & 1:0.16 & 1:0.7 \\
{Cs-rate} & 20\% & 25\% \\
\hline
\end{tabular} }
\end{table}
The main issue of training code-switching model is lack of adequate code-switching training sentences because code-switching mostly occurs in speech or personal messages instead of in written resources. We think that generating code-switching sentences from monolingual data may solve the above issue since we can obtain monolingual text much easier than code-switching text. In the following examples and discussion, Mandarin is the host language, and English is the guest language. Actually, the proposed approach is language independent, so it is possible to apply on other host-guest language pairs.

\begin{figure}[ht]
\includegraphics[width=\linewidth]{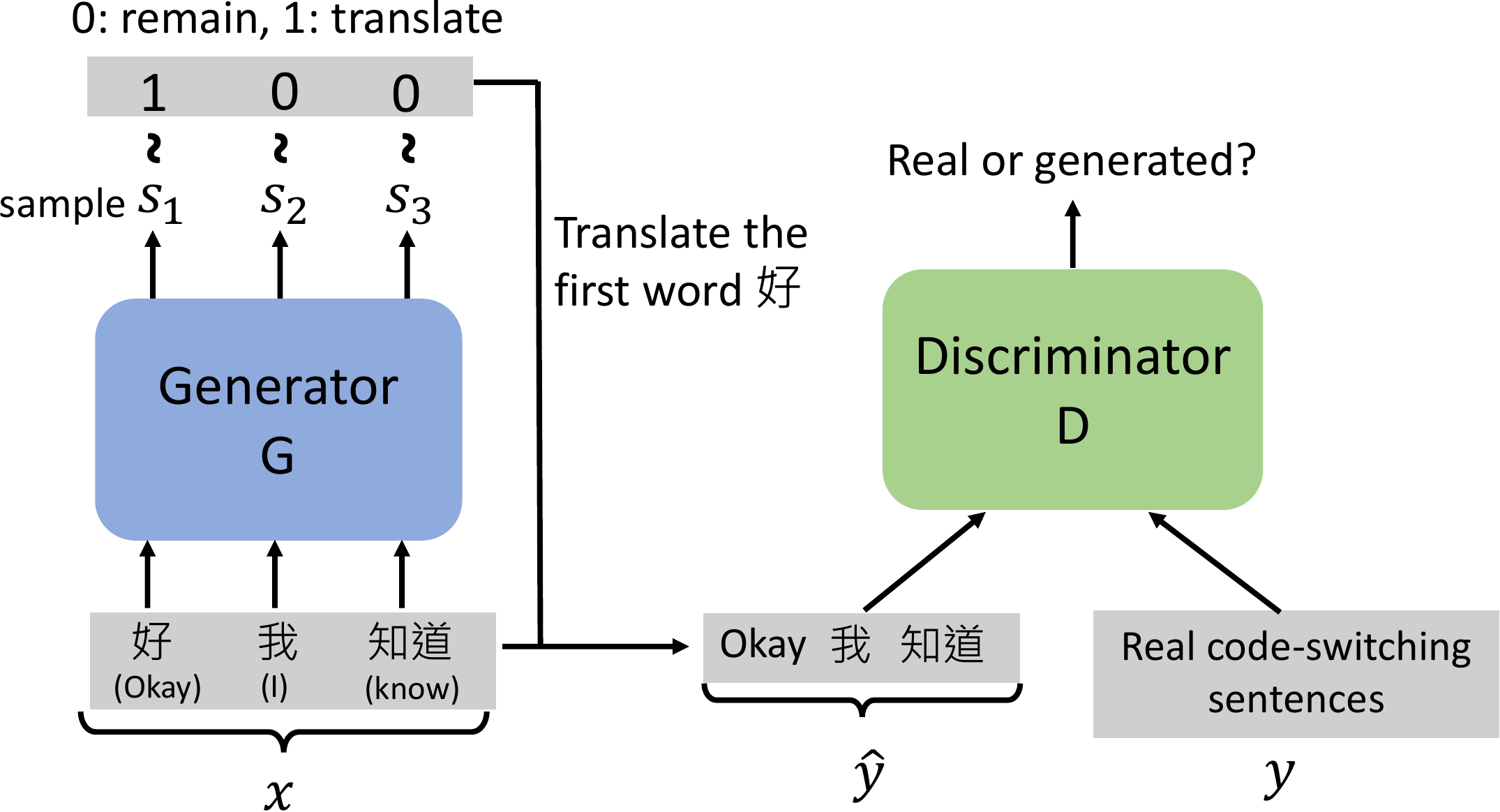}
\caption{\footnotesize Proposed framework. The generator learns to generate code-switching sentences from monolingual sentences.}
\label{fig:overall}
\end{figure}

To generate intra-sentential code-switching sentences, we can randomly select some of the words in the Mandarin sentence and translate them into English. However, this approach will generate many unreasonable code-switching sentences.
\begin{CJK*}{UTF8}{bkai}
For instance, given the monolingual sentence ``我要介紹 ... '' (I will introduce ...), replacing the  word ``我'' with ``I'' does not  generate a reasonable sentence because few speakers would code-switch in this way.
\end{CJK*}
Nevertheless, there are no perfect rules to predict which word or phrase in a sentence should be code-switched or not. Inspired by GAN, we propose to learn a conditional generator for code-switching sentences, so it can transform a monolingual sentence into a code-switching sentence.

\textbf{Discriminator}.
The discriminator $D$ takes a sentence as input, and outputs a scalar between 0 and 1. The output scalar indicates that the input sentence is generated by the generator G or from the given code-switching training sentence. For a well-trained perfect discriminator, the output is zero when the sentence is generated by generator G, and the output is one when the sentence is sampled from the training data set.

\textbf{Generator}.
The generator $G$ takes a monolingual (Chinese) word sequence $x=\{x_1,x_2,...,x_N\}$ as input, where $N$ is the length of $x$\footnote{As typical conditional GAN, the generator $G$ also takes a noise $z$ sampled from Gaussian as input. We ignore $z$ in the following formulation for simplicity.}. The output of $G$ is a sequence of values $s=\{s_1,s_2,...,s_N\}$. $s_n$ is a scalar between 0 and 1 corresponding to input word $x_n$. Each scalar $s_n$ represents whether it is proper to replace $x_n$ with its translated counterpart. $s_n$ is considered as a probability, and a binary value is sampled from it. If the sampled value is $1$, the word in Mandarin will be translated into English. By contrast, if $0$ is sampled, the input word will remain the same.
A code-switching sentence $\hat{y}$ is thus generated from $G$. The generator proposed here only learns which word in Mandarin can be replaced with English. It is possible to have a generator which directly generates code-switching word sequence. However, learning which word can be replaced is easier than learning to generate the words in another language directly.

\textbf{Training of Discriminator}.
$D$ is learned by minimizing $\mathcal{L}_D$ below,
\begin{equation}
\mathcal{L}_D = - ( \mathbb{E}_{y \sim \mathcal{D}_{cs}} [ log D(y) ] + \mathbb{E}_{x \sim \mathcal{D}_{zh}, \hat{y}\sim G(x)} [ log (1-D(\hat{y})) ] ).
\label{eq:d}
\end{equation}
In the first term of (\ref{eq:d}), the code-switching sentence $y$ is sampled from the training data $\mathcal{D}_{cs}$, and the discriminator $D$ learns to assign a larger score $D(y)$ to  $y$. In the second term, a monolingual sentence $x$ is sampled from a data set $\mathcal{D}_{zh}$, and $G$ transforms $x$ into a code-switching sentence $\hat{y}$. $D$ learns to assign a smaller score $D(\hat{y})$ to $\hat{y}$.

\textbf{Training of Generator}.
The parameters in $G$ are learned from the following loss function $\mathcal{L}_G$.
\begin{equation}
\mathcal{L}_G = - \mathbb{E}_{x \sim \mathcal{D}_{zh}, \hat{y}\sim G(x)} [ log D(\hat{y}) ].
\label{eq:g}
\end{equation}
With (\ref{eq:g}), $G$ learns to generate $\hat{y}$ that can obtain large $D(\hat{y})$. 
Due to the output of discriminator is discrete, the model is updated by the REINFORCE algorithm~\cite{williams1987class,williams1992simple, sutton2000policy}.

The discriminator $D$ and generator $G$ are trained iteratively as typical GAN.

\section{Experimental setup}
\label{sec:exp}
\vspace{-1mm}

\subsection{Corpora}
\label{ssec:corpora}
\vspace{-1mm}

In this work, we utilized two data sets for the experiments: LectureSS and SEAME corpus~\cite{lyu2010seame}. The detailed statistics of these corpora are listed in Table~\ref{table:corpus}. Additionally, we draw a comparison between them in Table~\ref{table:corpusvs}. CS-rate in Table~\ref{table:corpusvs} is defined as below,
\begin{equation}
\mbox{CS-rate}=\frac{\mbox{\# English words in CS utterances}}{\mbox{\# total words in
CS utterances}}.
\label{eq:cs_rate}
\end{equation}

LectureSS is a lecture speech corpus recorded by one Taiwanese instructor at National Taiwan University in 2006. The content of the recording is ``Signal and System" (SS) course. It is spontaneous speech with highly imbalanced Mandarin-English code-switching characteristics. Mandarin is the host language and English is the guest language. Most English words in this corpus are domain-specific terminologies.

South East Asia Mandarin-English (SEAME) corpus is a conversational speech corpus recorded by Singapore and Malaysia speakers with almost balanced gender in Nanyang Technological University
and Universities Sains Malaysia. There are two speaking types in the speech: conversational and interview conditions, and the content are related to daily life, school, and so on. It is also Mandarin-English code-switching while the amount of Chinese (ZH) words and English (EN) words is about equal. Not only proper nouns but also conjunctions may be used in English in this corpus.
Some sentences in SEAME are completely in English, while it does not happen in NTU lecture.
Nevertheless, the cs-rate of LectureSS is close to the cs-rate of SEAME.

\begin{table*}[ht]
\caption{\footnotesize Code-switching point (CSP) prediction on manually labeled sentences.}
\label{table:exp1}
\centering {\footnotesize
\begin{tabular}{l|ccc|c|ccc}
\hline
 & {Precision} & {Recall} & {F-measure}
   & {BLEU-1} & {WER(\%)} & {EN WER} & {ZH WER} \\
\hline \hline
 & \multicolumn{7}{c}{LectureSS} \\
\hline
{ZH}        & 0    & 0    & 0    & 0.76 & 20.56 & 100   & 0     \\
{EN}        & 0.21 & 1    & 0.35 & 0.20 & 102.1 & 0     & 128.5 \\
\hline
{random}    & 0.17 & 0.16 & 0.16 & 0.62 & 39.20 & 88.14 & 26.54 \\
{noun}      & \textbf{0.55} & 0.44 & 0.22 & 0.75 & \textbf{17.06} & 54.02 & \textbf{10.21} \\
\hline
{proposed}      & 0.52 & 0.42 & 0.46 & 0.78 & 22.82 & 54.24 & 14.69 \\
{proposed+pos}  & 0.52 & \textbf{0.55} & \textbf{0.53} & \textbf{0.80} & 21.08 & \textbf{39.83 } & 16.23 \\
\hline \hline
 & \multicolumn{7}{c}{SEAME} \\
\hline
{ZH}        & 0    & 0    & 0    & 0.65 & 33.51 & 100   & 0     \\
{EN}        & 0.3  & 1    & 0.46 & 0.01 & 80.35 & 0     & 117.3 \\
\hline
{random}    & 0.26 & 0.23 & 0.24 & 0.47 & 47.02 & 76.50 & 33.07 \\
{noun}      & \textbf{0.61} & 0.19 & 0.29 & 0.49 & 45.44 & 93.99 & 22.48 \\
\hline
{proposed}      & 0.55 & 0.35 & 0.43 & \textbf{0.58} & \textbf{30.00} & 61.20 & \textbf{15.25} \\
{proposed+pos}  & 0.51 & \textbf{0.47} & \textbf{0.49} & 0.52 & 33.33 & \textbf{48.63} & 26.10 \\
\hline
\end{tabular} }
\end{table*}

Before using these datasets, we cleaned them first. LectureSS comprises of ``Zhuyin fuhao", mathematical symbols and English alphabet which cannot be translated into English words or Chinese words. In addition, SEAME contains non-speech labels, unknown words labels, incomplete words and foreign words. We removed these words directly if the semantics of the sentences would not be influenced too much; otherwise, we ignored the utterances in the experiments.

\vspace{-1mm}
\subsection{Model Setup}
\label{ssec:setup}
\vspace{-1mm}

The inputs of both the discriminator and the generator are word sequences. There are two ways to represent a word. In the first approach, each word is first represented by one-hot encoding, and transforms into an embedding by an embedding layer. We set 8200 and 12000 vocabulary size for LectureSS and SEAME individually, and 150 as the dimension for word embedding. In the second approach, we also consider the part-of-speech (POS) tag for each word. We used Jieba\footnote{Jieba toolkit from: https://github.com/fxsjy/jieba}, an Open Source Chinese segmentation application in Python language, as our POS tagger. Only Chinese words are tagged and English words are tagged as ``eng.'' Each POS tag corresponds to a 64-dim one-hot encoding, and it is transformed into 20-dim by an embedding layer. The embedding of words and POS tags are concatenated. The embedding layer is jointly trained with the whole model, and Chinese and English word embedding are trained together.

The generator $G$ is made up of embedding layer, one bidirectional long short-term memory (BLSTM)~\cite{graves2005framewise} layer, one fully connected (FC) layer. It outputs one value with sigmoid for each time step to determine whether this word will be translated into English. Gaussian noise is 10-dim vector concatenated with the output of BLSTM. The parameter of $G$ is updated by policy gradient with the output of $D$ as reward. Translator is merely a mapping table which contains a list of Chinese vocabulary with each comparing English word translated by Google translator.

The discriminator $D$ shares the same embedding layer and BLSTM with $G$. However, it is updated only when $G$ is training and fixed when $D$ is training. The output of BLSTM is passed into a FC layer with dropout rate 0.3. It ends in a one-dimension vector with sigmoid.

The whole optimization process is based on Adam optimizer~\cite{kingma2014adam} and we train 100 epochs for all experiments. The input data of $G$ is all Chinese training sentences, and $D$ is trained by all code-switching sentences in the training set and fake code-switching sentences generated by $G$ in respective corpora.

\vspace{-1mm}
\section{Results}
\label{sec:result}

\begin{CJK*}{UTF8}{bkai}

\begin{table*}[ht]
\caption{\footnotesize Code-switching examples from different methods.}
\label{table:example}
\centering {\footnotesize
\begin{tabular}{c|l}
\hline

\hline
{Ground Truth} & {Causality 這個~也是~讀~過~的~就是~指~我~output at-any-time~只~depend-on input} \\\hline\hline
{Input}        & {因果性~這個~也是~讀~過~的~就是~指~我~輸出~在任意時間~只~取決於~輸入} \\
               & {(Causality, this is also what you have read, that means what I output
at any time only depends on input)} \\\hline
\hline
{Random}       & {因果性~this 也是~讀~過~的~就是~指~我~output~在任意時間~只~取決於~輸入} \\\hline
{Noun}         & {Causality~這個~也是~讀~過~的~就是~指~我~輸出~在任意時間~只~取決於~input} \\\hline
{Proposed}         & {Causality this also~你~所 read 過~的~就是~指~我~output~在任意時間~只~取決於~輸入} \\\hline
{Proposed+pos}     & {Causality~這個~也是~讀~過~的~就是~指~我~output at-any-time~只~depend-on~輸入} \\\hline
\end{tabular} }
\end{table*}

\end{CJK*}

\newcommand{\expOne}{Code-switching Point Prediction}
\newcommand{\expTwo}{Generated Text Quality}
\newcommand{\expThree}{Language Modeling}

We evaluate our proposed method in three aspects: code-switching point (CSP) prediction, quality of generated text, and performance of language modeling with augmented text.

\vspace{-1mm}
\subsection{\expOne}
\label{ssec:exp1}
\vspace{-1mm}

We selected 50 code-switching sentences $y$ in testing set as ground truth, and manually translated them into fully Chinese sentences $x$. The generator then generates code-switching sentences $\hat{y}$ conditioned on $x$. We consider the positions of English words in $y$ as CSPs that we want to detect, and use precision, recall and F-measure to evaluate the accuracy of detected CSPs in $\hat{y}$. Additionally, we also apply BLEU score~\cite{papineni2002bleu} and word error rate (WER) in (\ref{eq:wer}) to evaluate  $\hat{y}$,
\begin{equation}
\mbox{Word Error Rate}=\frac{\sum_i\mbox{Edit Distance}(y_i, \hat{y_i})}{\mbox{\# total words}},
\label{eq:wer}
\end{equation}
where $i$ indicates the $i^{th}$ selected code-switching sentences.

The proposed approach is compared with four baselines:
(1) \textit{ZH}: fully Chinese sentences, that is, $\hat{y} = x$.
(2) \textit{EN}: fully English sentences.
(3) \textit{random}: words are randomly translated into English. The translation probability for each word is the same as the cs-rate of the corpus considered.
(4) \textit{noun}: translate all words that are tagged as nouns (common nouns and proper nouns)~\cite{wei2006mixing} by POS tagger into English.

According to Table~\ref{table:exp1},  we observe that \textit{ZH} can have good performance in BLEU score and total WER. This is attributed to the fact that Chinese words occur more frequently than English words in code-switching sentences in both corpora. \textit{Random} gets poor performance because people don't code-switch arbitrarily.  \textit{Noun} has high precision owing to high exactness, but it does not predict CSPs other than noun.
Our method obtains better recall, F-measure, BLEU-1 and English WER than \textit{noun} on all the corpora because it detects not only nouns but other CSPs like conjunctions, discourse particles, filled pauses, and so on.

\vspace{-1mm}
\begin{table}[ht]
\caption{\footnotesize Quality of generated code-switching text from testing text evaluated by PPL on n-gram LM and neural-based LM (RNNLM).}
\label{table:exp2}
\centering {\footnotesize
\begin{tabular}{l|cc|cc}
\hline
 & {random} & {noun} & {proposed} & {proposed+pos}
\\
\hline \hline
\multicolumn{5}{c}{LectureSS} \\
\hline
{n-gram} & 1022.95 & 337.713 & \textbf{330.028} & 400.275 \\
{RNNLM} & 82.292 & 75.515 & 82.802 & \textbf{74.287} \\
\hline
\multicolumn{5}{c}{SEAME} \\
\hline
{n-gram} & 177.039 & 154.28  & 159.103 & \textbf{145.3} \\
{RNNLM} & 79.338 & 84.081 & 78.335 & \textbf{69.345} \\
\hline
\end{tabular} }
\end{table}

\vspace{-1mm}
\subsection{\expTwo}
\label{ssec:exp2}
\vspace{-1mm}

Next, we demonstrate the quality of our generated text by validating them with language model trained on training text. We calculate the perplexity (PPL) of our generated text.

Two types of language models were used to evaluate our results: n-gram model~\cite{brown1992class} and neural language model~\cite{bengio2003neural}. N-gram language model is a word-level tri-gram with Kneser-Ney (KN) smoothing~\cite{kneser1995improved} trained by SRILM~\cite{stolcke2002srilm}. Recurrent neural networks based language model (RNNLM) is a two-layer character-level LSTM~\cite{hochreiter1997long} language model. Because the two corpora have different scales of training data, we used 32-dimensional LSTM for LectureSS and 64-dimensional LSTM for SEAME. We used Adam optimizer with initial learning rate 0.5 to optimize the network, and 0.7 dropout is also applied to avoid over-fitting.

We used \textit{random}, \textit{noun} and the proposed approaches to generate some sentences to evaluate text quality.  These sentences are generated from all the fully Chinese sentences in the testing set (810 sentences for LectureSS and 2211 sentences for SEAME as shown in Table~\ref{table:corpus}). The results are shown in Table~\ref{table:exp2}.  In the result of both language models, we observe that the performance of our methods with POS tagging (\textit{proposed+pos}) is far better than \textit{random} on both corpora. It shows that our model has the capability to transform a Chinese sentence into a code-switching sentence with the similar pattern as the training data.

\vspace{-1mm}
\begin{table}[ht]
\caption{\footnotesize PPL of neural based language model (RNNLM) trained on code-switching training text and data augmented from Chinese training text. The last \textit{+pos} column indicates considering POS features in proposed method.}
\label{table:exp3}
\centering {\footnotesize
\begin{tabular}{l|l|c|cc|cc}
\hline
 & & {train} & {random} & {noun}
 & {proposed} & {+pos}
\\
\hline \hline
{LectureSS} & {dev} & 110.35 & 107.28 & 105.37 & 109.58 & \textbf{103.37} \\
\cline{2-7}
           & {test} & 73.394 & 71.779 & 70.038 & 71.974 & \textbf{69.185} \\
\hline\hline
{SEAME} & {dev} & 75.295 & 75.307 & 75.307 & \textbf{74.474} & 75.119 \\
\cline{2-7}
        & {test}& 86.088 & 83.873 & 85.366 & \textbf{83.819} & 84.358 \\
\hline
\end{tabular} }
\end{table}

\vspace{-1mm}
\subsection{\expThree}
\label{ssec:exp3}
\vspace{-1mm}

To see whether the data augmentation methods help language modeling, we trained RNNLM which is introduced in Section~\ref{ssec:exp2} on training data, and evaluated them on the same set of development data and testing data. We do not show the results of n-gram-based LM here because its performance is not comparable with RNNLM as shown in Table~\ref{table:exp2}. Lower perplexity represents better performance on language modeling. We form the augmented training set by combining the generated code-switching sentences with the original training set. The generated code-switching sentences are from the Chinese sentences in original training set (4643 sentences in LectureSS and 20365 sentences in SEAME as shown in Table~\ref{table:corpus}).

Table~\ref{table:exp3} shows our experimental results. The \textit{train} column in the table represents the perplexity of language model without the augmented code-switching sentences, which is the baseline of the experiment. As shown in this table, \textit{random} surpasses the baseline on both LectureSS and SEAME testing set. It shows that augmented code-switching text helps language modeling even if the CSPs are randomly selected. \textit{Noun} improves the results only on LectureSS. This may be due to the fact that LectureSS contains lots of CSPs on domain-specific noun, while SEAME has more complicated CSPs.

\textit{Proposed+pos} performs the best on LectureSS, while \textit{proposed} performs the best on SEAME\footnote{We notice that the influence of POS tags in Table~\ref{table:exp3} and Table~\ref{table:exp2} are different. However, we believe the results in Table~\ref{table:exp3} is more crucial because it is the goal of this task. In Table~\ref{table:exp2}, even the model decides not to code-switch any word in the input sentences, it may still obtains reasonable number, but this model would not be very helpful in Table~\ref{table:exp3}.}. It indicates that POS features help generator generate more useful code-switching sentences on LectureSS, but not SEAME. This is because in LectureSS domain-specific terminologies which tend to be code-switched into English are nouns. Meanwhile, SEAME comes from daily life conversation where CSPs are not focused on nouns, resulting in better performance without POS information. Based on Table~\ref{table:exp3}, we demonstrated that the augmented data automatically generated by our method helps language modeling on CS text, and by adding POS features to generator input, our generated data further improves RNNLM on some of the data domains.

\vspace{-1mm}
\subsection{Examples}
\label{ssec:obs}
\vspace{-1mm}

Some generated code-switching examples are demonstrated in Table~\ref{table:example}. The first row is the original code-switching sentence (\textit{ground truth}). We translated it into fully Chinese (\textit{input}). Then, we compared the generated results to \textit{random} and \textit{noun}. The rule-based approach is accurate, but  cannot find out all CSPs. The proposed method with POS tagging can find out more CSPs. More examples are in the following link: \url{http://goo.gl/KdBYSy}. The examples show that the proposed approach usually generates reasonable code-switching sentences.  However, it also generates some terrible sentences.
We found that most of them stem from bad translation from Chinese to English.

\section{Conclusion}
\label{sec:concl}

In this work, we try to generate code-switching sentences from monolingual sentences by GAN. The generator can learn to predict CSPs to a great degree without any linguistic knowledge. Moreover, our generated code-switching sentences are better than random generation and rule-based generation. Last but not least, the augmented data by our methods improves RNNLM. For the future work, there is still room for improvement in the translator in this work since wrong translation may lead to terrible generated code-switching sentences. We will further analyze the generator to learn more mechanism about code-switching.

\bibliographystyle{IEEEtran}
\bibliography{mybib}

\end{document}